\documentclass[conference]{IEEEtran}
\pdfoutput=1
\usepackage{times}
\usepackage{amsmath}
\usepackage[numbers,sort,compress]{natbib}
\usepackage{multicol}
\usepackage[bookmarks=true]{hyperref}
\usepackage{amssymb}
\usepackage[linesnumbered]{algorithm2e}
\usepackage{booktabs}
\usepackage{multirow}
\usepackage{graphicx}
\usepackage{bmpsize}
\usepackage{siunitx}
\usepackage{float}
\usepackage[font=small,skip=10pt]{caption}
\usepackage{stackengine}

\begin{document}

\title{DICP: Doppler Iterative Closest Point Algorithm}

\ifdefined\RSSSubmission{} 
\author{Author Names Omitted for Anonymous Review. Paper-ID 50}
\else
\author{
Bruno Hexsel, Heethesh Vhavle and Yi Chen\\
Corresponding email: \href{mailto:research@aeva.ai}{research@aeva.ai} \\
Aeva, Inc, Mountain View, CA 94043
}
\fi

\maketitle

\begin{abstract}

In this paper, we present a novel algorithm for point cloud registration for range sensors capable of measuring per-return instantaneous radial velocity: Doppler ICP. Existing variants of ICP that solely rely on geometry or other features generally fail to estimate the motion of the sensor correctly in scenarios that have non-distinctive features and/or repetitive geometric structures such as hallways, tunnels, highways, and bridges.
We propose a new Doppler velocity objective function that exploits the compatibility of each point's Doppler measurement and the sensor's current motion estimate.
We jointly optimize the Doppler velocity objective function and the geometric objective function which sufficiently constrains the point cloud alignment problem even in feature-denied environments.
Furthermore, the correspondence matches used for the alignment are improved by pruning away the  points from dynamic targets which generally degrade the ICP solution.
We evaluate our method on data collected from real sensors and from simulation. Our results show that with the added Doppler velocity residual terms, our method achieves a significant improvement in registration accuracy along with faster convergence, on average, when compared to classical point-to-plane ICP that solely relies on geometric residuals.

\end{abstract}

\IEEEpeerreviewmaketitle

\section{Introduction}

\begin{figure}
 \center
  \includegraphics[width=\columnwidth]{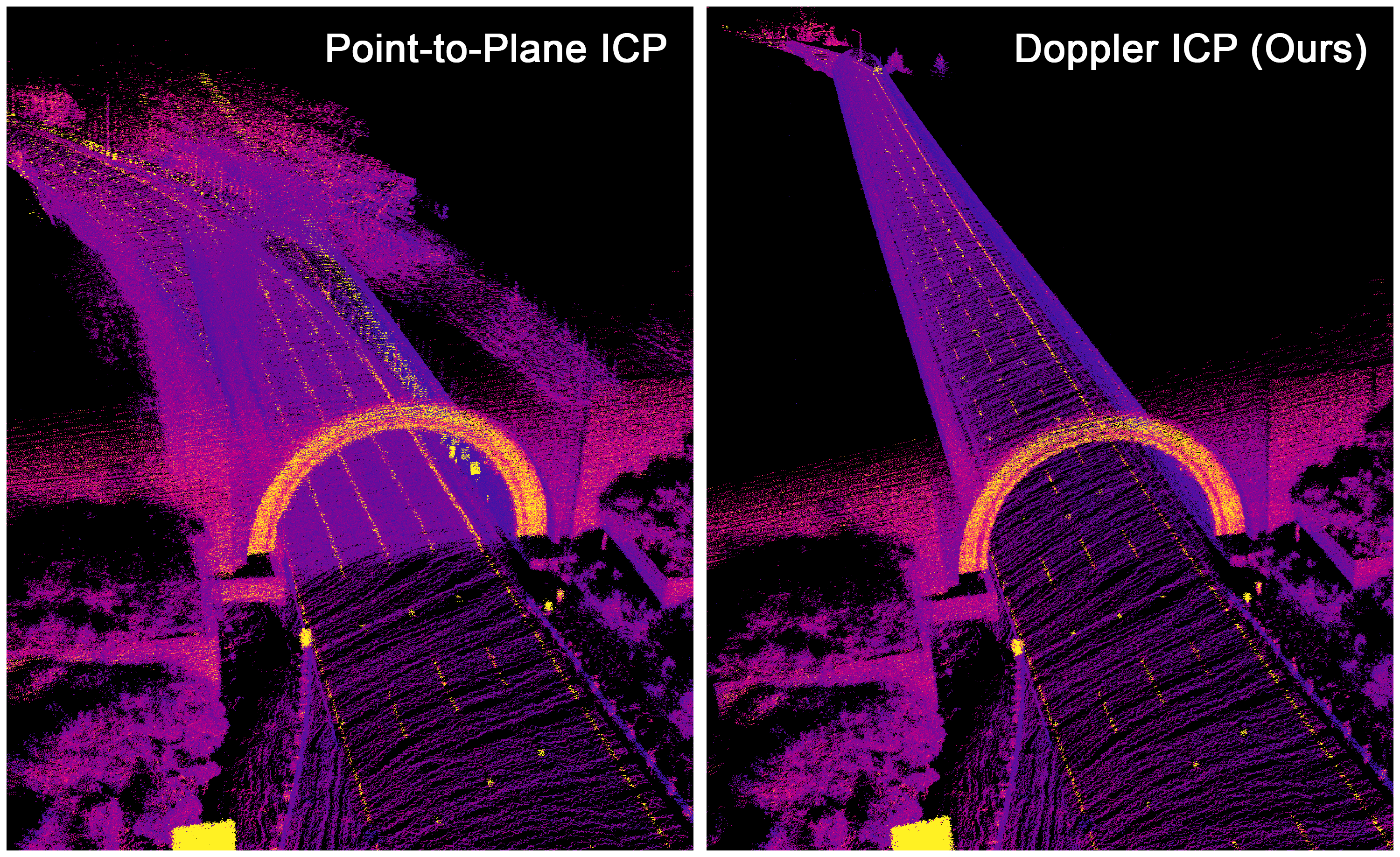}
  \caption{Comparison of tunnel reconstructions using point-to-plane ICP (left) and Doppler ICP (right) with measurements collected by an FMCW LiDAR.
  Point-to-plane ICP fails in this degenerate case due to the lack of distinctive features in the scene whereas the Doppler ICP algorithm is able to reconstruct the scene with very low error.}
  \label{tunnel}
\end{figure}

The problem of geometrically aligning two point clouds, also known as point cloud registration,
has many applications in robotics, healthcare, and others \cite{bellekens2014survey}.
In the past decades, methods for solving point cloud registration have been extensively researched and many algorithms have been
developed for this purpose \cite{bellekens2015benchmark}. Among those, one popular algorithm is Iterative Closest Point (ICP) \cite{chen1992object}.
In ICP, two point clouds, defined as the source and the target point clouds, are matched by iteratively
finding a transform that minimizes predefined distance metric between them. Generally speaking, registration methods based on ICP can lead to high-precision results but remain vulnerable in the presence of geometrically non-distinctive or repetitive environments, such as tunnels, highways, or bridges \cite{barsan2020learning}, as shown in Figure \ref{tunnel} (left).

Despite these shortcomings, most rangefinder sensors that generate a point cloud can make use of the ICP algorithm for point cloud registration.
For instance, RGB-D sensors provide measurements of images synchronized with depth images and can use the
point cloud generated from the depth image along with information from the image for point cloud registration \cite{kim2013image}.
Light detection and ranging (LiDAR) sensors are another class of rangefinder sensors with the broad use of ICP methods.
These sensors use coherent light to measure the bearing, range, and often intensity of return points.
One promising development in LiDAR technology in recent years is the advent of coherent frequency-modulated continuous-wave (FMCW)
LiDARs, which employs direct modulation and demodulation of the laser waveform in frequency domain \cite{pierrottet2008linear}.
Compared to two other popular LiDAR schemes of pulsed and amplitude-modulated continuous-wave (AMCW) \cite{behroozpour2017lidar}, FMCW LiDARs measure the beat frequency, a frequency difference resulting from an alternating constructive and destructive interference pattern caused by the outgoing and incoming signal, to indirectly measure the range of the target \cite{royo2019overview}. This scheme also allows FMCW LiDARs to be able to measure the relative velocity between each measured point to the sensor along the radial direction (Doppler velocity) by the Doppler effect, therefore providing additional motion information about the target other than range and intensity that other types
of LiDARs cannot support.
Furthermore, the FMCW technology can also be applied to radars to allow them
to measure the Doppler velocity \cite{vivet2013localization}.

The advancement of the FMCW technology in range sensors offers a new dimension of information in the raw measurement level for each measured point, creating opportunities for traditional methodologies that only rely on range information to be revisited. For instance, the Doppler velocity measurement could be used to tackle the aforementioned feature-denied challenging environments for ICP algorithms. However, to our best knowledge, no such work has been reported to explore this possibility and we aim to fulfill this gap with our proposed method. 

In this paper, we present a novel Doppler ICP (DICP) algorithm that takes into consideration the measured Doppler velocity for each measured point by FMCW LiDARs or radar sensors in the registration process. In DICP, we first establish the relationship between the Doppler velocity of an observed point and the transformation to be estimated. A Doppler velocity residual is then defined and included in the overall objective function to be minimized jointly with the classical geometric objectives used in ICP.
The method is validated using both real-world data collected using an FMCW LiDAR on a driving vehicle and simulated data using the CARLA simulator \cite{dosovitskiy2017carla}. 

The key contribution of our work can be summarized as: 
\begin{itemize}
    \item By deriving the Doppler error distance metrics and their gradients, we show that virtually any variant of ICP can be extended to include these terms to improve registration accuracy and robustness.
    \item We demonstrate that, by using gradients derived from Doppler velocity measurements, the overall objective function incorporates terms that are independent of the structure of the environment, which is particularly important for the optimization to converge with point clouds captured in environments that are challenging to most variants of ICP as discussed above.
    \item The convergence rate of ICP  is greatly improved by introducing the Doppler error component and gradient terms in addition to geometric error and gradient terms.
    \item We show that correspondence matches used for the optimization step in the registration algorithm are improved by exploiting the compatibility of each point’s Doppler measurement and the current sensor’s rigid motion estimate. This enhances the algorithm performance by eliminating most moving objects in a scene that would otherwise degrade the performance of point cloud registration.
\end{itemize}

\section{Related Work}
\label{related}

Several enhancements to the ICP algorithm have been proposed with different
performance profiles in real-world datasets \cite{bellekens2014survey}.
In \cite{serafin_grisetti_2014}, the augmentation of ICP is discussed
by using normals and tangents to the surface as part of the optimization step.
A point-to-plane algorithm is combined with the classical ICP algorithm \cite{segal2009generalized}.In \cite{park2017colored}, the authors jointly optimize a photometric objective function with a geometric objective function; the approach taken in this paper is similar to the approach described in \cite{park2017colored} where instead we jointly optimize a geometric objective function with a Doppler-derived objective function.

We discuss a formulation where the linear and angular components
are considered to derive the 3D rigid transformations used in the computation of ICP.
A similar formulation to this paper can be found in \cite{hong2010vicp}, where
the authors use the sensor velocity components to
estimate the distortion effect from the sensor motion while building the point cloud.

Our proposed algorithm uses a formulation for instantaneous radial velocity
measured as Doppler velocity to apply to a point cloud matching algorithm.
In the experiments section, this measured point cloud is obtained by 
making use of an FMCW LiDAR.
In the context of research for using Doppler enabled sensors for positioning,
\cite{vivet2013localization} presents an approach where a rotating FMCW radar is used to
estimate ego-motion and build a map based on the static returns.
In \cite{sim2020road} and \cite{de2020kradar++}, the FMCW
radar spectral information is used for place and pose estimation.
\cite{guo2021doppler} proposed a Doppler velocity-based cluster and velocity estimation algorithm using an FMCW LiDAR. 

One of the issues addressed in our work, outlier rejection of dynamic objects in a scene,
is addressed in \cite{rodriguez2007improved,kim2013image,chetverikov2002robust}.
Note that the algorithm described in this paper takes advantage of the Doppler velocity
measurements for outlier rejection independently of robust outlier rejection via any
other method involving a robust loss functions.
In practice, DICP can be combined with robust loss methods to further enhance the performance of the point cloud registration algorithm.

\section{Method}
\label{method}

\subsection{Notation}

We denote $\mathcal{F}_{I}$ to be the inertial frame, $\mathcal{F}_{V}$ to be the vehicle body frame,  and $\mathcal{F}_{L}$ to be the LiDAR body frame.
For convenience, we also denote the vehicle body frame for the previous point cloud as the
\emph{source frame}, $\mathcal{F}_S$, and that for the current point cloud as the \emph{target frame}, $\mathcal{F}_T$.

Generally, we denote ${}_{F}\textbf{t}_{AB} \in \mathbb{R}^3$ as a vector from point $A$ to point $B$  expressed in the frame $\mathcal{F}_F$. In this configuration, the transformation matrix $\textbf{T}_{AB} \in SE(3)$ transforms a vector ${}_{B}\textbf{t}_{AB}$ expressed in $\mathcal{F}_{B}$ to the frame $\mathcal{F}_{A}$. $\mathbf{R}_{AB} \in SO(3)$ rotates a vector expressed in frame $\mathcal{F}_B$ to frame
$\mathcal{F}_A$.

We denote the source point cloud $\mathcal{P}$ as a set of points such that $P_j$ is the $j$-th point and $p_j$ is the corresponding tuple of the point containing at least the 3D vector ${}_{S}\mathbf{t}_{SP_j}$ (which
is the vector from the origin $\mathcal{F}_S$ to a source point $P_j$ expressed in the source frame) and the measured Doppler
velocity $v_{meas_{j}}$.
For the target point cloud $\mathcal{Q}$, we denote
$Q_j$ to be the $j$-th point and $q_j$ to be the corresponding point tuple containing the 3D vector ${}_{T}\mathbf{t}_{TQ_j}$
representing a vector from the origin $\mathcal{F}_T$ to the target point $Q_j$ expressed in the target frame. The tuple $q_j$ may also contain
other relevant information such as the surface normal at the point $Q_j$ denoted as
${}_{T}\mathbf{n}_{Q_j}$.

Last but not least, $\hat{\mathbf{a}}$ is denoted as the skew-symmetric matrix of vector $\mathbf{a}$, and a homogeneous coordinate of a vector $\textbf{a} \in \mathbb{R}^3$ is defined as
$\bar{\textbf{a}} = \begin{bmatrix}\textbf{a}^\top & 1 \end{bmatrix}^\top$.

\subsection{A Primer on ICP and State Vector Representation}
The ICP algorithm aims to estimate a rigid-body transformation, $\textbf{T}_{TS}$ that best aligns 
the source point cloud $\mathcal{P}$ to
the target point cloud $\mathcal{Q}$. The general procedure of ICP can be described as correspondence association, error minimization, and outlier filtering. In the first stage, the correspondence between points is established using the Euclidean distance \cite{besl1992method}, geometric features \cite{chen1992object} or colors of points \cite{park2017colored}. Next, $\textbf{T}_{TS}$ is estimated by iteratively minimizing
\begin{equation}\label{eq_argmin_T}
    \textbf{T}^*_{TS} = \arg\min_{\textbf{T}_{TS}}
      E\left(\textbf{T}_{TS}, \mathcal{P}, \mathcal{Q}\right),
\end{equation}
where $E\left(\textbf{T}_{TS}, \mathcal{P}, \mathcal{Q}\right)$ is the error function for all matched correspondences between point cloud $\mathcal{P}$ and
 $\mathcal{Q}$ with all points in $\mathcal{P}$ transformed using $\textbf{T}_{TS}$.
After the transformation is estimated and applied, the process is repeated by removing outliers and redefining correspondences. This process is
iterated until either the solution converges or the termination criteria is reached, usually when the overall error function stops evolving significantly.
One widely adopted ICP variant is the point-to-point ICP algorithm, where the error function is defined based on the Euclidean distance between correspondences:
\begin{equation}\label{def_point_to_point_residual}
    E\left(\textbf{T}_{TS}, \mathcal{P}, \mathcal{Q}\right) =
    \sum_{j=1}^{N}\|\textbf{T}_{TS} {}_{S}\bar{\textbf{t}}_{SP_j} - {}_{T}\bar{\textbf{t}}_{TQ_j}\|^2,
\end{equation}
where $P_j$ and $Q_j$ are a pair of correspondences, $j \in [1, ..., N]$ and $N$ represents the number of pairs of correspondence. Note that for the simplicity of notation, the L2 norm operator $\|\cdot\|^2$ in Equation \ref{def_point_to_point_residual} and the dot product operator $\cdot$ in Equation \ref{def_point_to_plane_residual} only apply to the first three coefficients of the homogeneous coordinates.

Another type of algorithm is the point-to-plane ICP algorithm, where the residual is defined as the projection of Euclidean distance onto the surface normal at the corresponding target point: 
\begin{equation}\label{def_point_to_plane_residual}
    E\left(\textbf{T}_{TS}, \mathcal{P}, \mathcal{Q}\right) = 
    \sum_{j=1}^{N}\left( (\textbf{T}_{TS} {}_{S}\bar{\textbf{t}}_{SP_j} - {}_{T}\bar{\textbf{t}}_{TQ_j})\cdot {}_{T}\bar{\textbf{n}}_{Q_j} \right)^2,
\end{equation}

To minimize the error function, either a non-linear least square method such as the Gauss-Newton method and Levenberg-Marquardt algorithm \cite{fitzgibbon2003robust} or Iteratively Reweighted Least-Squares (IRLS) with robust kernels can be applied \cite{bergstrom2014robust}. To do that, we first denote a 6-dimensional coordinate $\mathbf{u}$ in the Lie algebra $\mathfrak{se}(3)$ space to represent the state vector:
\begin{equation}\label{def_u}
\mathbf{u} = \begin{bmatrix} \mathbf{u}_{\theta}^\top & \mathbf{u}_{t}^\top \end{bmatrix} ^ \top,
\end{equation}
where $\mathbf{u}_{\theta}$ is the rotation component such that $\mathbf{u}_{\theta} = [\text{Log}(\mathbf{R}_{TS})]^\vee$ and $\mathbf{u}_{t} = {}_{T}\mathbf{t}_{TS}$ is the translation component. Then we denote the pseudo-exponential map of $\mathbf{u}$ as in \cite{blanco2010tutorial}, such that 
\begin{equation}\label{eq_T_TS}
    \textbf{T}_{TS} = \text{pseudo-exp}(\mathbf{u}) =
    \begin{bmatrix}
      e^{\mathbf{\hat{u}}_\theta} & \mathbf{u}_t \\
      \mathbf{0} & 1
    \end{bmatrix}.
\end{equation}
Note $\mathbf{R}_{TS} = e^{\mathbf{\hat{u}}_\theta}$ is the exact exponential map, whereas the $\mathbf{u}_{t}$ is left intact during the mapping. By doing so, it defines a valid retraction on $SE(3)$ and also leads to a more efficient computation of the Jacobians. Thus, the minimization problem described by Equation \ref{eq_argmin_T} becomes
\begin{equation}\label{eq_argmin_u}
    \mathbf{u}^* = \arg \min_{\mathbf{u}} E(\mathbf{u}, \mathcal{P}, \mathcal{Q}).
\end{equation}

In the following subsections, we will derive the relation between Doppler velocity and the state vector $\mathbf{u}$, which will be used to formulate new residual terms as well as in the outlier rejection in the DICP algorithm. 

\subsection{Doppler Velocity}\label{ss:doppler-velocity}

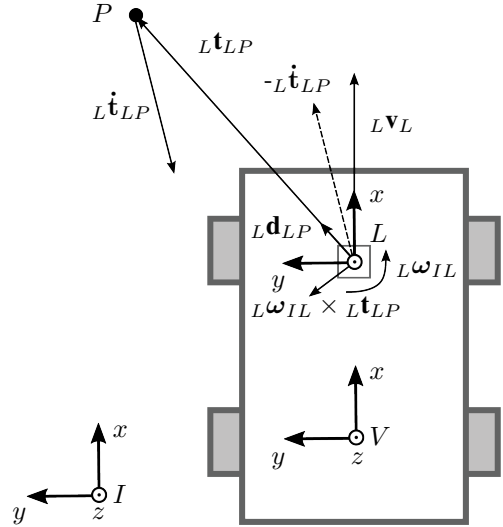
\begin{figure}[h]
    \begin{picture}(150,200)
        \put(20,0){\includegraphics[scale=0.1]{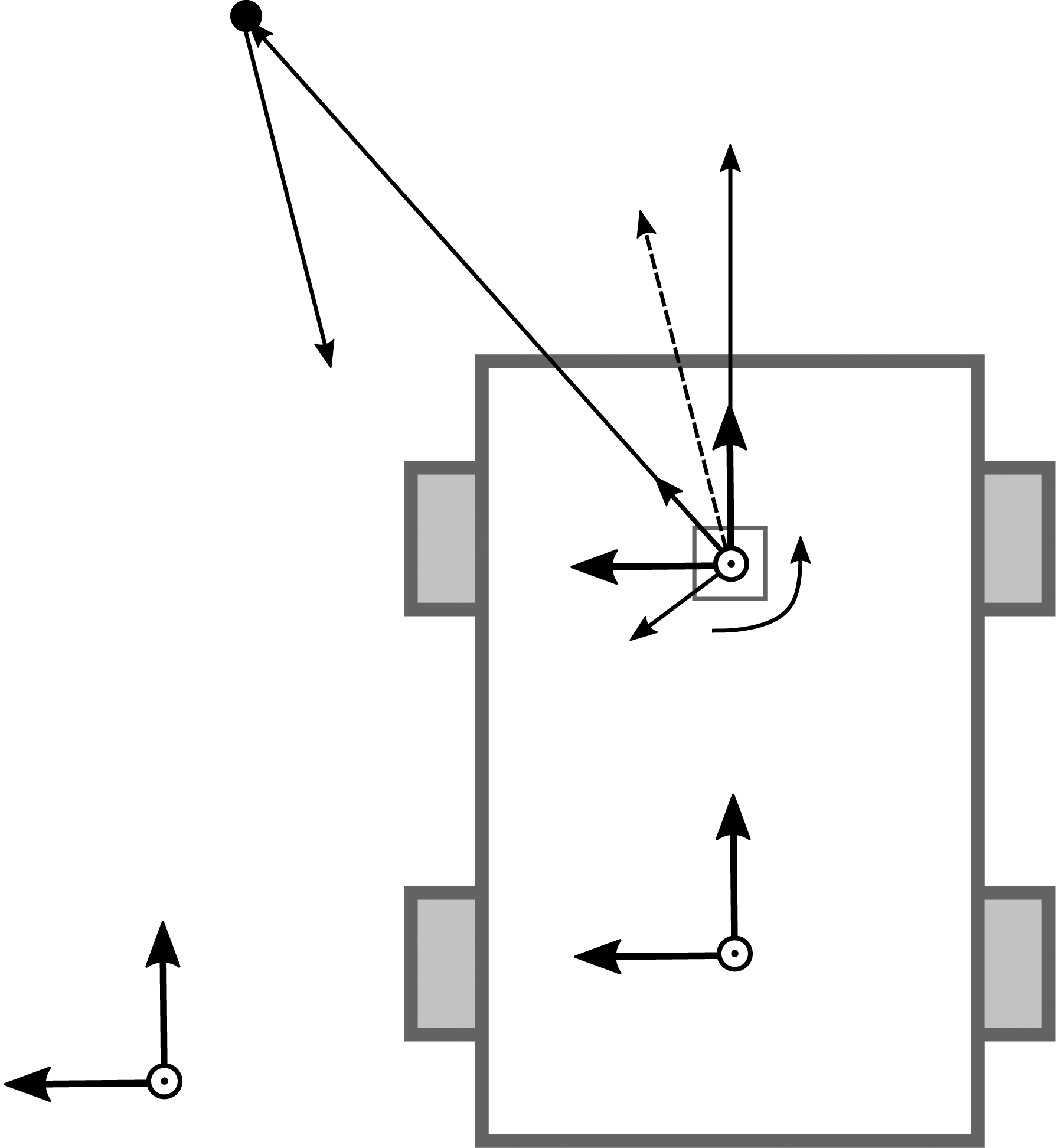}}
        \put(150, 107){$L$}
        \put(113, 90){$y$}
        \put(150, 122){$x$}
        \put(150, 30){$V$}
        \put(113, 23){$y$}
        \put(143, 23){$z$}
        \put(150, 55){$x$}
        \put(53, 8){$I$}
        \put(15, 2){$y$}
        \put(45, 2){$z$}
        \put(53, 33){$x$}
        \put(45, 190){$P$}
        \put(85, 180){${}_{L}\textbf{t}_{LP}$}
        \put(105, 110){${}_{L}\textbf{d}_{LP}$}
        \put(150, 150){${}_{L}\textbf{v}_{L}$}
        \put(160, 95){${}_{L}\boldsymbol\omega_{IL}$}
        \put(105, 80){${}_{L}\boldsymbol\omega_{IL} \times {}_{L}\textbf{t}_{LP}$}
        \put(45, 155){${}_{L}\mathbf{\dot{t}}_{LP}$}
        \put(110, 165){-${}_{L}\mathbf{\dot{t}}_{LP}$}
    \end{picture}
    \caption{An illustration of reference frames, position, and velocity vectors. The velocity of a
    static point $P$ with respect to the origin of $\mathcal{F}_{L}$, ${}_{L}\mathbf{\dot{t}}_{LP}$,
    is the resultant vector of
    ${}_{L}\textbf{v}_{L}  + {}_{L}\boldsymbol\omega_{IL} \times {}_{L}\textbf{t}_{LP}$
    (shown as the dashed arrow) with an opposite sign.}
    \label{fig:frames}
\end{figure}

It is assumed that the LiDAR\footnote{
In this and the following sections, the name LiDAR is used to refer to a range measurement
device that can measure Doppler velocity for each point. Even though we
use the term LiDAR to refer to this class of sensors, it should
be noted that it is not restricted to laser-only devices and this methodology can be generalized to a broader array of range sensors that are capable of measuring Doppler velocity, such as FMCW radars.
} is mounted on a rigid body on the
vehicle and the offsets ${}_{V}\mathbf{t}_{VL}$ and rotation
$\mathbf{R}_{VL}$ (the LiDAR sensor extrinsic calibration), are known by calibration beforehand.

We also denote the vehicle's velocity expressed in the vehicle frame as ${}_{V}\textbf{v}_{V} \in \mathbb{R}^3$ and the angular velocity to be ${}_{V}\boldsymbol\omega_{IV} \in \mathbb{R}^3$. The velocity of the LiDAR expressed in the vehicle frame is defined as ${}_{V}\textbf{v}_{L}$ and the angular velocity to be ${}_{V}\boldsymbol\omega_{IL}$. From the aforementioned
assumption that the LiDAR is rigidly mounted on the vehicle, the LiDAR
velocity in the vehicle frame is given by
\begin{equation}\label{eq_Vs}
    {}_{V}\textbf{v}_{L} = {}_{V}\textbf{v}_{V} + {}_{V}\boldsymbol\omega_{IV} \times {}_{V}\textbf{t}_{VL}.
\end{equation}

As illustrated in Figure \ref{fig:frames}, suppose a stationary point $P$ exists within the detectable region of the LiDAR sensor, then its relative velocity ${}_{L}\mathbf{\dot{t}}_{LP}$ expressed in the LiDAR frame $\mathcal{F}_{L}$ can be expressed as
\begin{equation}
    {}_{L}\textbf{v}_{P} = {}_{L}\textbf{v}_{L} + {}_{L}\mathbf{\dot{t}}_{LP} + {}_{L}\boldsymbol\omega_{IL}  \times {}_{L}\textbf{t}_{LP},
\end{equation}
where ${}_{L}\textbf{t}_{LP}$ is the range measurement of the LiDAR,
${}_{L}\textbf{v}_{L}$ is the LiDAR velocity expressed in $\mathcal{F}_{L}$.
Since the point $P$ is stationary with respect to the inertial frame, it follows that ${}_{L}\textbf{v}_{P} = \textbf{0}$. We then have the relative velocity of point $P$ with respect to the LiDAR frame, ${}_{L}\mathbf{\dot{t}}_{LP}$ expressed as
\begin{equation}\label{rel-doppler}
    {}_{L}\mathbf{\dot{t}}_{LP} = -{}_{L}\textbf{v}_{L} - {}_{L}\boldsymbol\omega_{IL} \times {}_{L}\textbf{t}_{LP}.
\end{equation}
This derivation is illustrated in Figure \ref{fig:frames} as well.

\begin{figure}
 \center
  \includegraphics[width=\columnwidth]{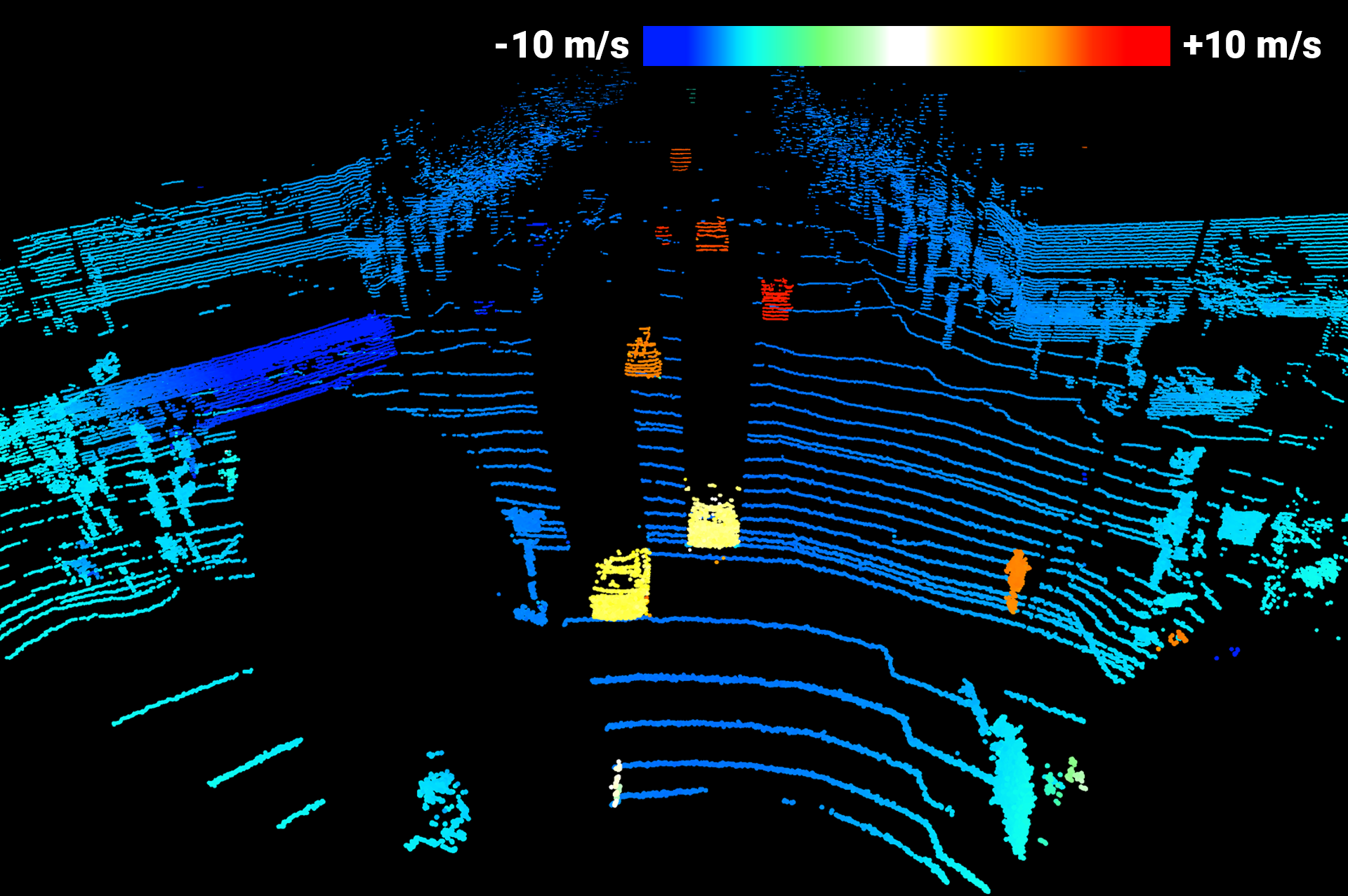}
  \caption{Point cloud colored based on the relative Doppler velocity measured by an FMCW LiDAR sensor. All points moving away from the sensor are gradually colored red and points moving towards the sensor are gradually colored blue. The static points in this scene appear blue due to the negative relative Doppler velocity observed on them as the ego-vehicle moves forward.}
  \label{fig:doppler-pcd}
\end{figure}

Doppler sensors only measure the relative radial velocity,
i.e. the relative velocity projected along the direction vector between the sensor and the point. The direction vector from the LiDAR to the point ${}_{L}\textbf{d}_{LP}$ is expressed as
\begin{equation}
    {}_{L}\textbf{d}_{LP} = \frac{{}_{L}\textbf{t}_{LP}}{\parallel {}_{L}\textbf{t}_{LP}\parallel}.
\end{equation}

We here define the Doppler velocity as the projection of the
relative velocity vector ${}_{L}\mathbf{\dot{t}}_{LP}$ along
${}_{L}\textbf{d}_{LP}$  as ${}_{L}v_{LP}\in\mathbb{R}$ and we
further expand it using the definition from Equation \ref{rel-doppler}:
\begin{equation}
\begin{split}
    {}_{L}v_{LP} & = {}_{L}\textbf{d}_{LP} \cdot {}_{L}\mathbf{\dot{t}}_{LP} \\
    & = -{}_{L}\textbf{d}_{LP} \cdot ({}_{L}\textbf{v}_{L} + {}_{L}\boldsymbol\omega_{IL} \times {}_{L}\textbf{t}_{LP}).
\end{split}
\end{equation}

Note that vector ${}_{L}\textbf{d}_{LP}$ has the same direction of the reading vector ${}_{L}\textbf{t}_{LP}$ and is therefore orthogonal to the resultant vector of the cross product of ${}_{L}\textbf{t}_{LP} \times {}_{L}\boldsymbol\omega_{IL}$. Hence, it follows that ${}_{L}\textbf{d}_{LP} \cdot  ({}_{L}\textbf{t}_{LP} \times {}_{L}\boldsymbol{\omega}_{L}) = \textbf{0}$, thus $ {}_{L}v_{LP}$ is given by
\begin{equation}\label{eq_v_do}
     {}_{L}v_{LP} = -{}_{L}\textbf{d}_{LP} \cdot {}_{L}\textbf{v}_{L}.
\end{equation}

To express ${}_{L}v_{LP}$ in the vehicle frame, we substitute Equation \ref{eq_v_do} into Equation \ref{eq_Vs} and obtain
\begin{equation}\label{eq_v_sp_in_V}
\begin{split}
    {}_{L}v_{LP} & =-{}_{L}\textbf{d}^{\top}_{LP} \left(\mathbf{R}_{LV} {}_{V}\textbf{v}_{L} \right) \\
    & =-\left({}_{L}\textbf{d}^{\top}_{LP} \mathbf{R}^{\top}_{VL} \right) {}_{V}\textbf{v}_{L}  \\
    & =-\left(\mathbf{R}_{VL} {}_{L}\textbf{d}_{LP}  \right)^{\top} {}_{V}\textbf{v}_{L},
\end{split}
\end{equation}
which can be expressed, with the direction vector ${}_{L}\textbf{d}_{LP}$, in the vehicle 
frame $\mathcal{F}_V$ as
\begin{equation}\label{eq_dv}
    {}_{V}\textbf{d}_{LP} = \mathbf{R}_{VL} {}_{L}\textbf{d}_{LP}.
\end{equation}
Substituting Equations \ref{eq_Vs} and \ref{eq_dv} into Equation \ref{eq_v_sp_in_V}, we obtain the Doppler velocity
component expressed in the vehicle frame
\begin{equation}\label{doppler-in-v}
    {}_{L}v_{LP} = -{}_{V}\textbf{d}_{LP} \cdot ({}_{V}\textbf{v}_{V} - {}_{V}\mathbf{\hat t}_{VL} {}_{V}\boldsymbol\omega_{IV}).
\end{equation}

Notice that this is the general definition where the vehicle frame
$\mathcal{F}_V$ does not coincide with the LiDAR frame
$\mathcal{F}_L$. If the vehicle frame is chosen to be the same
as the LiDAR frame, the angular velocity ${}_{V}\boldsymbol\omega_{IV}$ is no longer measurable
as a Doppler velocity component.
This is because Doppler measurement devices only measure
radial velocity components and cannot directly observe tangential velocity
components elicited by rotation. Furthermore, the Doppler velocity measurements are also independent of the geometry of the observed objects. For instance, the Doppler measurements of a smooth wall are not affected by the fact that the wall is featureless, as only the relative radial velocity between each point and the sensor matters. Figure \ref{fig:doppler-pcd} shows a point cloud colored by the relative Doppler velocity measured by a FMCW LiDAR.

\subsection{Doppler Velocity Residual}\label{ss:doppler-residual}

To incorporate the Doppler velocity into the formulation of the ICP algorithm, we first approximate the angular velocity
${}_{V}\boldsymbol\omega_{IV}$ within the period of one LiDAR sample $\Delta t$ as
\begin{equation}\label{eq_omega_to_phi}
  {}_{V}\boldsymbol\omega_{IV} \approx -\frac{\mathbf{u}_{\theta}}{\Delta t}.
\end{equation}
Similarly, the linear velocity ${}_{V}\textbf{v}_{V}$ is
approximated as the amount of translation within the period of one LiDAR sample $\Delta t$ as
\begin{equation}\label{eq_v_to_t}
  {}_{V}\textbf{v}_{V} \approx -\frac{\mathbf{u}_{t}}{\Delta t}.
\end{equation}

The negative signs in Equations \ref{eq_omega_to_phi} and \ref{eq_v_to_t} are present because the state vector that we are optimizing corresponds to the transform $\textbf{T}_{TS}$, whereas the linear and angular velocities are approximated time derivatives of the transform $\textbf{T}_{ST}$, which represents the vehicle's motion from the source to the target frame. 

Furthermore, to simplify the derivation of the Jacobians of the corresponding residual terms, both definitions above use the time derivative of Lie algebra components to approximate the velocities, assuming that the rotation of the transformation between the source and target point cloud is relatively small and the sampling rate is fast enough.

For a given state-vector $\mathbf{u}$, the residual between the measured Doppler
velocity and the expected Doppler velocity for the $j$-th point
$P_{j}$ given a state-vector $\mathbf{u}$ is given by
\begin{equation}\label{doppler-error}
r_{v_j} = v_{meas_j} -{}_{L}v_{LP_j}(\mathbf{u}).
\end{equation}
Substituting Equation \ref{eq_omega_to_phi} and \ref{eq_v_to_t} into Equation \ref{doppler-in-v}, we have the Doppler velocity for $P_{j}$ expressed as
\begin{equation}\label{doppler-vel-with-Lie}
    {}_{L}v_{LP_{j}}(\mathbf{u}) = \frac{1}{\Delta t} {}_{V}\textbf{d}_{LP_j} \cdot
\left( \mathbf{u}_{t} - {}_{V}\mathbf{\hat t}_{VL} \mathbf{u}_{\theta} \right).
\end{equation}
Substituting Equation \ref{doppler-vel-with-Lie} into Equation \ref{doppler-error}, the residual of Doppler velocity term becomes:
\begin{equation}\label{eq_doppler_error}
r_{v_j} = v_{meas_j} -\frac{1}{\Delta t} {}_{V}\textbf{d}_{LP_j} \cdot
\left( \mathbf{u}_{t} - {}_{V}\mathbf{\hat t}_{VL} \mathbf{u}_{\theta} \right).
\end{equation}

Equation \ref{eq_argmin_u} can be solved using a general-purpose non-linear optimization approach or IRLS with robust kernels. In order to do so, the Jacobian of the residual with respect to the state is often required and we define the Doppler Jacobian term for the point $P$ as
\begin{equation}\label{jac-doppler}
    \mathbf{J}_{v_j} = \frac{\partial r_{v_j}}{\partial \mathbf{u}}= -\frac{1}{\Delta t} {}_{V}\textbf{d}^{\top}_{LP_j} \frac {\partial  \left( \mathbf{u}_{t} -
    {}_{V}\mathbf{\hat t}_{VL} \mathbf{u}_{\theta} \right)}{\partial \mathbf{u}}.
\end{equation}
Solving Equation \ref{jac-doppler}, we have
the solution for the Jacobian of the Doppler velocity residual of the $j$-th point expressed in frame $\mathcal{F}_V$:
\begin{equation}\label{jac-doppler-solved}
\mathbf{J}_{v_j} = \frac{1}{\Delta t}
\begin{bmatrix}
({}_{V}\textbf{d}_{LP_j} \times {}_{V}\mathbf{t}_{VL})^{\top} &
-{}_{V}\textbf{d}^{\top}_{LP_j} \end{bmatrix}.
\end{equation}

One important thing to notice is that in the above derivation the Doppler velocity residual term does not
depend on the target point cloud, as the Doppler velocity measurements are intrinsic to each point cloud and only those of the source point
cloud are used in our method. While it is possible for one to perform ICP steps by using the Doppler velocity residuals only, the algorithm
benefits greatly from combining with other types of geometric residuals such as point-to-point or
point-to-plane.
The reason is that, although the Doppler velocity residual terms are not correlated to the geometric relation between source and target point clouds, their gradients provide guidance for the overall objective to be minimized when
combined with other geometric residual terms.
Therefore, incorporating the Doppler velocity residuals and their gradients greatly
improve the performance of ICP when ambiguous structures are present in the point clouds,
such as long planar surfaces along hallways, tunnels, parking lots, etc.

It is also worth noting that since the target point cloud is not required to contain Doppler
velocity measurements, it means that this algorithm can 
be utilized not only when we are matching two sequential point clouds 
but can also be used to register a scan to a prior map of the world.

\subsection{Dynamic Point Outlier Rejection}\label{ss:outlier-rejection}

While a perfectly static point cloud registration is not affected by moving objects,
any registration where there are moving objects in a scene, can introduce errors
in the estimated registration.
Here we describe the use of the Doppler velocity measurements to reject moving objects
that are considered outliers.

For a given state-vector as described in Equation \ref{def_u}, it is possible to estimate
what the expected Doppler measurement should be and measure its error as defined
in Equation \ref{doppler-error}. Dynamic points from moving objects will be rejected if
their Doppler measurements don't agree with the expected Doppler velocity for
a given state-vector during the iteration of the algorithm.

Let $\Delta_{v}$ be the threshold that specifies the maximum deviation
from the expected Doppler velocity given the state-vector $\mathbf{u}$ and
the actual measured Doppler velocity for a given source point $P_j$, a point can be
determined to be an inlier if the following is true:

\begin{equation}\label{eq_dynamic_threshold}
\mid r_{v_j}\mid  < \Delta_{v}.
\end{equation}

Notice that $\Delta_{v}$ does not necessarily need to be a constant. It could
progressively become smaller as the algorithm converges as to not initially
reject valid static points due to a state-vector being initially far from the
optimal value.

\subsection{Doppler Iterative Closest Point Algorithm}

We wish to progressively apply a transformation to the source point cloud $\mathcal{P}$ to minimize an error function with respect to the target point cloud $\mathcal{Q}$, and we choose to solve the error function using the IRLS approach \cite{bergstrom2014robust}. In such a formulation, the error function is defined as the weighted sum of squared residuals.

Therefore, the Doppler error term $E_v(\mathbf{u}, \mathcal{P})$ can be defined as
\begin{equation}\label{eq_residual_p2p}
    E_v(\mathbf{u}, \mathcal{P}) =
    \sum^{N}_{j = 1}
    {w_v(r_{v_j})r_{v_j}^2},
\end{equation}
where $N$ is the number of correspondences, $j$ is the index of a correspondence pair, and $w_v(\cdot)$ is the weight function for the Doppler velocity residual component. The corresponding Jacobian with respect to the state vector $\mathbf{u}$ is shown in Equation \ref{jac-doppler-solved}.

In this work, we choose the point-to-plane metric to minimize the geometric error between the correspondence pairs, however, other distance metrics are also applicable here. The point-to-plane residual term is
defined as
\begin{equation}
    r_{p_j} =
    \left( \mathbf{R}_{TS} (\mathbf{u}_{\theta}) {}_{S}\mathbf{t}_{SP_j} + \mathbf{u}_{t}  - {}_{T}\mathbf{t}_{TQ_{j}} \right) \cdot {}_{T}\mathbf{n}_{Q_j}.
\end{equation}
We make the small-angle approximation, similar to \cite{behley2018efficient}, where $\mathbf{R} (\mathbf{u}_{\theta}) \approx I_3 + \hat{\mathbf{u}}_{\theta}$.
The Jacobian of the point-to-plane residual with respect to the state vector $\mathbf{u}$ for the $j$-th point is
\begin{equation}\label{eq_jac_p2p_icp}
    \mathbf{J}_{p_j} = \frac{\partial r_{p_j}}{\partial \mathbf{u}}=
      \begin{bmatrix}
      \left({}_{S}\mathbf{t}_{SP_j} \times {}_{T}\mathbf{n}_{Q_j}\right)^{\top} & {}_{T}\mathbf{n}_{Q_j}^{\top}
  \end{bmatrix}.
\end{equation}
The overall point-to-plane error function is expressed as
\begin{equation}\label{eq_p2p_icp}
    E_{p}\left( \mathbf{u}, \mathcal{P}, \mathcal{Q} \right) =
    \sum^{N}_{j = 1}
    {w_p(r_{p_j})r_{p_j}^2},
\end{equation}
where the term $w_p(\cdot)$ is the weight function for point-to-plane residual which could be a robust kernel.
We refer the reader to \cite{handa2014simplified} for
more details on point-to-plane ICP optimization.

The joint optimization of the Doppler velocity objective and the geometric objective can be weighed by a parameter $\lambda_{v}$, which remains constant over all the correspondences and is defined as
\begin{equation}\label{eq_full_opt}
    E(\mathbf{u}, \mathcal{P}, \mathcal{Q}) =
    \lambda_{v} E_v(\mathbf{u}, \mathcal{P})+
    (1 - \lambda_{v}) E_p(\mathbf{u}, \mathcal{P}, \mathcal{Q}).
\end{equation}
The objective function in Equation \ref{eq_full_opt} can be solved using an IRLS approach. The complete DICP algorithm along with dynamic point outlier rejection is summarized in Algorithm \ref{alg:dicp}.

\RestyleAlgo{ruled}
\begin{algorithm}[hbt!]
\caption{Doppler Iterative Closest Point}\label{alg:dicp}
\KwIn{
    \\
     $\mathcal{P}$ Source point cloud containing Doppler velocity measurements \\
     $\mathcal{Q}$ Target point cloud (optionally containing normals) \\
     $\Delta_d$ Maximum correspondence distance \\
     $\Delta_v$ Maximum velocity error \\
     $\mathbf{u}_0$ Initial state vector \\
     }
\KwOut{ \\
$\mathbf{u}$ Transform that aligns $\mathcal{P}$ with $\mathcal{Q}$
\\
}

$\mathbf{u} \leftarrow \mathbf{u}_0$

\While{not converged}{
    $\mathcal{P}^{\prime} \leftarrow \varnothing$

    $\mathcal{Q}^{\prime} \leftarrow \varnothing$

    \For{$j \leftarrow 1$ \KwTo $\mid \mathcal{P} \mid$ } {
        $p^{\prime}_{j} \leftarrow$ \texttt{TransformSourcePoint}($\mathbf{u}, p_j$)

        $q_{j} \leftarrow$ \texttt{FindClosestTargetPoint}($\mathcal{Q}, p^{\prime}_j$)

        \If {$\mid r_{v_j}\mid  < \Delta_{v} \wedge$
            \texttt{\upshape Dist}$(p^{\prime}_{j}, q_{j}) < \Delta_d $}{
            $\mathcal{P}^{\prime} \leftarrow \mathcal{P}^{\prime} \cup p_{j}$ \\
            $\mathcal{Q}^{\prime} \leftarrow \mathcal{Q}^{\prime} \cup q_{j}$
        }
    }

    $\mathbf{u} \leftarrow  \arg \min_{\mathbf{u}} E \left( \mathbf{u}, \mathcal{P}^{\prime}, \mathcal{Q}^{\prime} \right)$

}

\end{algorithm}

\section{Experiments}
\label{experiments}
In this section, we provide an overview of the dataset, discuss the different parameters used in the implementation, and show quantitative and qualitative results on several sequences which include sequential Doppler LiDAR scans from both real and simulated data.

\subsection{Dataset (Sequences)}

\begin{figure}[!t]
 \center
  \includegraphics[width=\columnwidth]{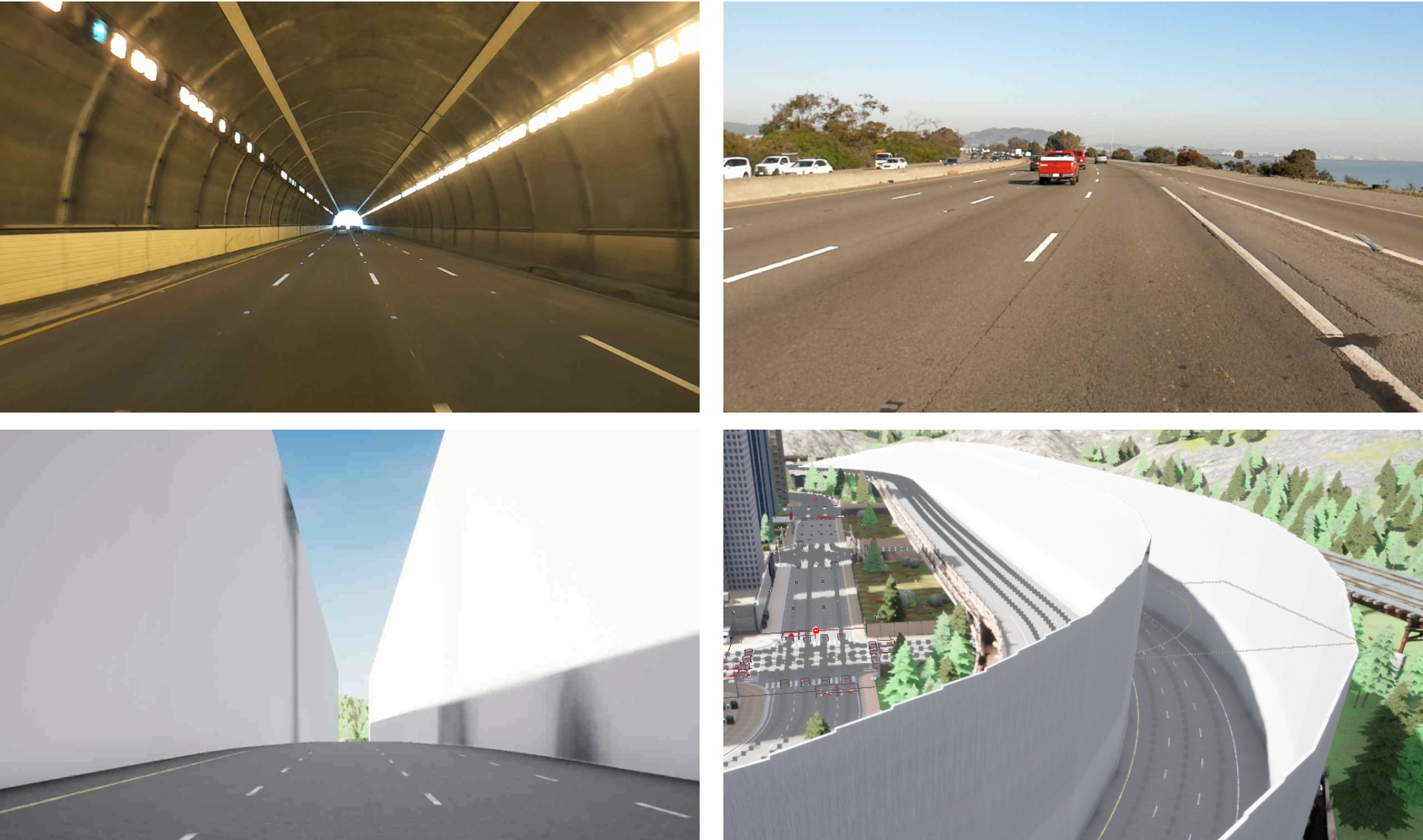}
  \caption{Representative scenes in the dataset. Robin Williams Tunnel in Sausalito, CA (top left); the stretch of US-101 near Brisbane Lagoon (top right); straight parallel walls placed in CARLA Town04 (bottom left); and curved walls placed in CARLA Town05 (bottom right) to simulate a feature-less scenario.}
  \label{fig:dataset}
\end{figure}

We evaluate our method on seven different sequences, of which five were collected from a real sensor and two were generated from a simulator. Since we are interested in understanding the effectiveness of Doppler ICP in feature-denied environments, most of the sequences include tunnels, highways, and bridges as shown in Figure \ref{fig:dataset}. As an exception, we also include an urban-driving scenario from San Francisco city which is rich in geometric features to benchmark our method where conventional ICP methods perform well. The key statistics of all sequences are provided in Table \ref{tab:dataset}. 

For the real-world data, we use Aeva's Aeries I \cite{AevaAeries1} FMCW LiDAR sensor to obtain the range and Doppler velocity measurements. The LiDAR sensor has a maximum horizontal field-of-view of $120^{\circ}$, a maximum vertical field-of-view of $30^{\circ}$, a 300 $\si{m}$ maximum operating range with precision of 2 \si{\cm} (one standard deviation), a Doppler velocity measurement precision of 3 \si[per-mode=symbol]{\cm\per\second} (one standard deviation), and a sampling rate of 10 \si{Hz}. Novatel ProPak6 GNSS \cite{novetal2015propak6} was used to obtain the ground truth odometry for these sequences. To generate the simulated sequences, we simulated an FMCW LiDAR
\footnote{The source code for our implementation of FMCW LiDAR is available in our fork of CARLA at \texttt{\href{https://github.com/aevainc/carla}{https://github.com/aevainc/carla}}.}
in the CARLA simulator based on the equations presented in Section \ref{ss:doppler-velocity}. We placed large parallel walls on either side of a straight highway and a curved highway, as shown in Figure \ref{fig:dataset}, to evaluate the effectiveness of our algorithm in a perfectly feature-denied environment. The ground truth transform was obtained by querying the CARLA actor's pose. The same scan pattern and sample rate from the Aeries I LiDAR were used in the simulation to resemble the behavior of the real LiDAR.

\begin{table}[!t]
\caption{\label{tab:dataset}Statistics of the dataset.}
    \centering
    \begin{tabular}{lrrr}
        \toprule
        \multicolumn{1}{c}{\multirow{2}{*}{Sequence}} & 
        \multicolumn{1}{c}{\multirow{2}{*}{\shortstack{Trajectory \\Length (m)}}} &
        \multicolumn{1}{c}{\multirow{2}{*}{\shortstack{Duration \\(seconds)}}} &
        \multicolumn{1}{c}{\multirow{2}{*}{\shortstack{\# Avg. \\Points}}} \\
        {} \\
        \midrule
Baker-Barry Tunnel (Empty)
& 860.31  %
& 83.7  %
& 55.7k \\  %

Baker-Barry Tunnel (Vehicles)
& 906.86  %
& 65.5  %
& 36.5k \\  %

Robin Williams Tunnel
& 688.76  %
& 30.0  %
& 43.3k \\  %

Brisbane Lagoon Freeway
& 4941.80  %
& 176.3  %
& 40.0k \\  %

San Francisco City
& 1378.14  %
& 450.0  %
& 120.1k \\   %

CARLA Town04 (Straight Walls)
& 599.91  %
& 46.4  %
& 78.8k \\  %

CARLA Town05 (Curved Walls)
& 426.81  %
& 76.0  %
& 80.4k \\  %

        \bottomrule
    \end{tabular}
\end{table}
    
\begin{table*}[h]
\caption{\label{tab:rmse-main}Point cloud registration results.}
    \centering
    \begin{tabular}{llrrrrrrrr}
        \toprule
        \multicolumn{1}{c}{\multirow{3}{*}{Sequence}} &
        \multicolumn{1}{c}{\multirow{3}{*}{Method}} &
        \multicolumn{4}{c}{No Seed Estimate} &
        \multicolumn{4}{c}{With Seed Estimate} \\
        \cmidrule(lr){3-6} \cmidrule(lr){7-10}
        {} & {} & \multicolumn{1}{c}{\shortstack{RPE\\Trans (m)}} & \multicolumn{1}{c}{\shortstack{RPE\\Rot (deg)}} & \multicolumn{1}{c}{\shortstack{Path\\Error (m)}} & \multicolumn{1}{c}{\shortstack{\# Iters\\(mean)}} & \multicolumn{1}{c}{\shortstack{RPE\\Trans (m)}} & \multicolumn{1}{c}{\shortstack{RPE\\Rot (deg)}} & \multicolumn{1}{c}{\shortstack{Path\\Error (m)}} & \multicolumn{1}{c}{\shortstack{\# Iters\\(mean)}} \\
        \midrule
\multirow{2}{*}{Baker-Barry Tunnel (Empty)}
& ICP-Open3D
& {1.0416}  %
& {0.1180}  %
& {525.35}  %
& {30.8}  %

& {0.7608}  %
& {0.1131}  %
& {203.86}  %
& {15.3} \\  %

& DICP (Ours)
& \textbf{0.0694}  %
& \textbf{0.1099}  %
& \textbf{1.23}  %
& \textbf{7.6}  %

& \textbf{0.0694}  %
& \textbf{0.1091}  %
& \textbf{1.24}  %
& \textbf{5.4} \\  %

\midrule

\multirow{2}{*}{Baker-Barry Tunnel (Vehicles)}
& ICP-Open3D
& {1.2641}  %
& {0.1817}  %
& {656.57}  %
& {26.1}  %

& {0.9385}  %
& \textbf{0.1468}  %
& {326.12}  %
& {14.6} \\  %

& DICP (Ours)
& \textbf{0.0807}  %
& \textbf{0.1493}  %
& \textbf{15.61}  %
& \textbf{8.4}  %

& \textbf{0.0807}  %
& {0.1498}  %
& \textbf{15.60}  %
& \textbf{6.1} \\  %

\midrule

\multirow{2}{*}{Robin Williams Tunnel}
& ICP-Open3D
& {1.8174}  %
& {0.1955}  %
& {366.84}  %
& {44.3}  %

& {1.9343}  %
& {0.1533}  %
& {219.10}  %
& {20.3} \\  %

& DICP (Ours)
& \textbf{0.0752}  %
& \textbf{0.1428}  %
& \textbf{0.03}  %
& \textbf{13.1}  %

& \textbf{0.0758}  %
& \textbf{0.1415}  %
& \textbf{0.05}  %
& \textbf{9.8} \\  %

\midrule

\multirow{2}{*}{Brisbane Lagoon Freeway}
& ICP-Open3D
& {2.7743}  %
& {0.2019}  %
& {4337.18}  %
& {37.5}  %

& {0.3270}  %
& {0.0915}  %
& {45.27}  %
& {10.6} \\  %

& DICP (Ours)
& \textbf{0.1132}  %
& \textbf{0.0894}  %
& \textbf{4.16}  %
& \textbf{17.2}  %

& \textbf{0.1301}  %
& \textbf{0.0869}  %
& \textbf{7.20}  %
& \textbf{9.7} \\  %

\midrule

\multirow{2}{*}{San Francisco City}
& ICP-Open3D
& {0.1853}  %
& {0.0510}  %
& {181.27}  %
& {11.8}  %

& {0.0323}  %
& \textbf{0.0479}  %
& {23.78}  %
& \textbf{5.9} \\  %

& DICP (Ours)
& \textbf{0.0308}  %
& \textbf{0.0489}  %
& \textbf{10.74}  %
& \textbf{7.4}  %

& \textbf{0.0317}  %
& {0.0482}  %
& \textbf{7.30}  %
& {6.0} \\  %

\midrule

\multirow{2}{*}{CARLA Town04 (Straight Walls)}
& ICP-Open3D
& {1.3935}  %
& {0.0313}  %
& {520.53}  %
& {28.5}  %

& {18.0178}  %
& {0.3004}  %
& {6446.52}  %
& {35.7} \\  %

& DICP (Ours)
& \textbf{0.0101}  %
& \textbf{0.0108}  %
& \textbf{0.40}  %
& \textbf{4.2}  %

& \textbf{0.0101}  %
& \textbf{0.0108}  %
& \textbf{0.41}  %
& \textbf{3.2} \\  %

\midrule

\multirow{2}{*}{CARLA Town05 (Curved Walls)}
& ICP-Open3D
& {0.2626}  %
& {0.0449}  %
& {100.29}  %
& {15.3}  %

& {0.2557}  %
& {0.0461}  %
& {91.60}  %
& {13.9} \\  %

& DICP (Ours)
& \textbf{0.0117}  %
& \textbf{0.0335}  %
& \textbf{1.50}  %
& \textbf{4.6}  %

& \textbf{0.0119}  %
& \textbf{0.0340}  %
& \textbf{1.51}  %
& \textbf{4.3} \\  %

        \bottomrule
    \end{tabular}
\end{table*}

\begin{figure*}[h]
 \center
  \includegraphics[width=\textwidth]{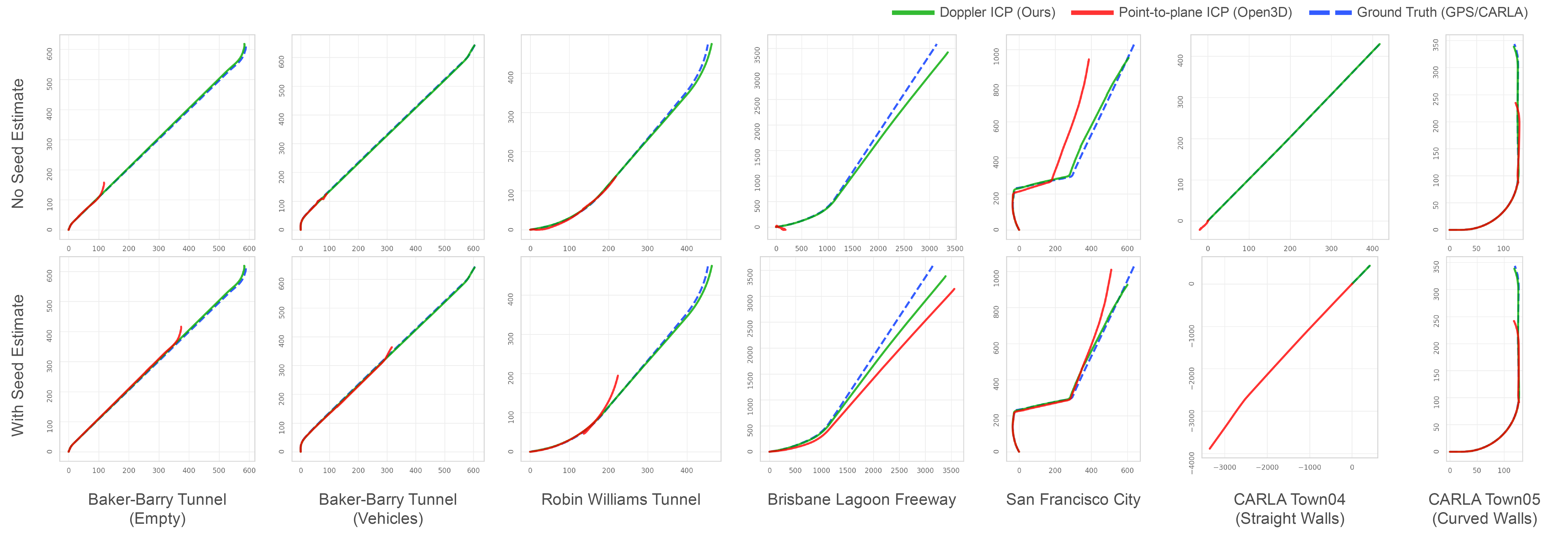}
  \caption{Trajectories for all sequences from composition of poses estimated using Doppler ICP (green lines) and Open3D point-to-plane ICP (red lines). The ground truth trajectories (based on GPS or obtained from CARLA simulation) are shown in dashed blue lines.}
  \label{fig-traj}
\end{figure*}

\subsection{Implementation}\label{ss:implementation}
We implement the Doppler ICP algorithm
\footnote{The code for Doppler ICP and our fork of Open3D is available at \texttt{\href{https://github.com/aevainc/Doppler-ICP}{https://github.com/aevainc/Doppler-ICP}}.}
by extending the existing point-to-plane implementation in Open3D \cite{Zhou2018}.
We employ robust kernels to minimize the effect of outlier correspondences (shown in Equations \ref{eq_residual_p2p} and \ref{eq_p2p_icp}) with Tukey loss with $k=0.5$ for the point-to-plane residual and Tukey loss with $k=0.2$ for the Doppler residual term. The Doppler residual robust kernel is enabled only after the second iteration of ICP to not reject too many correspondences in the initial few iterations where the error between the predicted and measured Doppler velocity would be high.
We set $\lambda_{v}=0.01$ in all our experiments, which was determined empirically. The dynamic point outlier rejection threshold from Equation \ref{eq_dynamic_threshold} is set to $\Delta_{v}=2 \si[per-mode=symbol]{\meter\per\second}$.

We separate our evaluation into groups of \textit{No Seed Estimate} and \textit{With Seed Estimate}. For \textit{No Seed Estimate} experiments, we do not initialize or seed the pose estimate for the registration methods. For \textit{With Seed Estimate} experiments, we use the estimated pose from the previous pair of registered scans in the sequence under a constant-velocity motion-model assumption. This is a more practical use case for ICP. We do not rely on any external sensor data and solely rely on the range and Doppler velocity measurements from the FMCW LiDAR when seeding the pose estimate. Other ways to seed the pose estimate include using the Doppler velocity measurements to estimate the ego-vehicle velocity and consequently estimate the pose in the sample period, or using additional sensors such as an IMU (Inertial Measurement Unit) along with pre-integration as proposed in \cite{Forster-RSS-15}. 

\begin{figure*}
 \center
  \includegraphics[width=\textwidth]{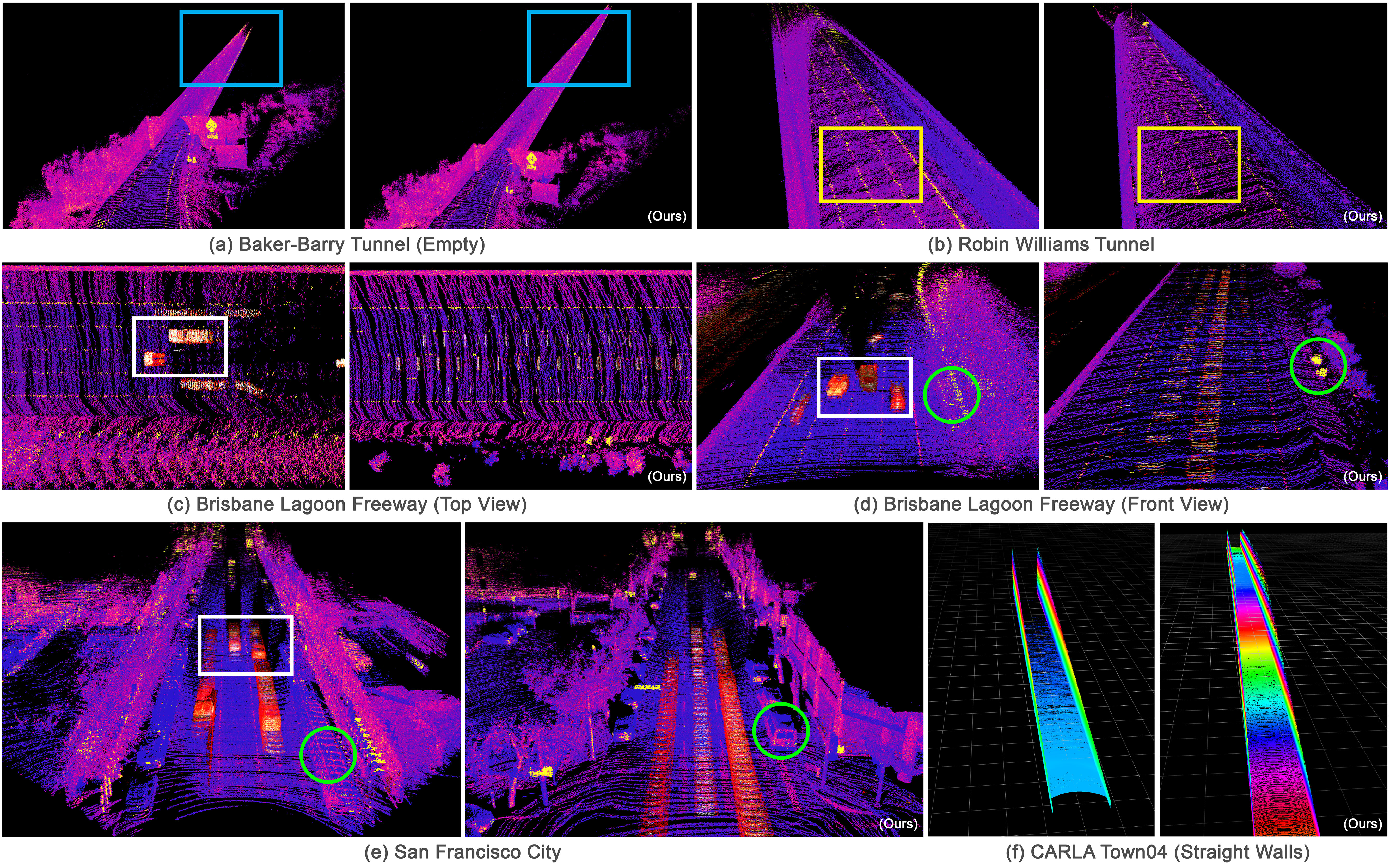}
  \caption{Reconstruction of different sequences by the composition of pose estimates. In each case, the left figure represents Open3D's point-to-plane ICP results and the right; Doppler ICP (ours). We only use the reflectivity channel from the LiDAR for visualization purposes in (a) to (e) and the height colormap in (f). (a), (b), and (f) demonstrate that point-to-plane ICP fails to reconstruct the entire length of the tunnel(s) correctly due to insufficient geometric constraints. In (a), the blue box highlights DICP's success in reconstructing the entire tunnel. The yellow boxes in (b) show how the lane markers can be clearly differentiated in our method. The white boxes in (c), (d), and (e) show that point-to-plane ICP registers to the moving vehicles in the scene instead of the static points. The green circles here highlight the resultant registration artifacts with point-to-plane ICP.}
  \label{fig:reconstructions}
\end{figure*}

\subsection{Results and Discussion}
We benchmark our performance (labeled \texttt{DICP}) against Open3D's point-to-plane ICP (labeled \texttt{ICP-Open3D}) with a robust kernel (Tukey loss, $k=0.5$) and do not use any Doppler velocity measurements there. All methods are evaluated based on the Relative Pose Error (RPE) metric \cite{8758020} in translation and rotation between the ground-truth and the estimated relative pose using point cloud registration. We also report the error in the total length of the estimated trajectory and the number of iterations each registration method took to converge. 

The metrics are presented in Table \ref{tab:rmse-main} and the estimated trajectories are shown in Figure \ref{fig-traj}. Figures \ref{tunnel} and \ref{fig:reconstructions} show a comparison of the scenes reconstructed using different registration methods. The trajectories are purely composed of sequential pose estimates represented in the inertial frame and no filtering or graph-based optimization techniques are used.

\begin{figure}
 \center
  \includegraphics[width=\columnwidth]{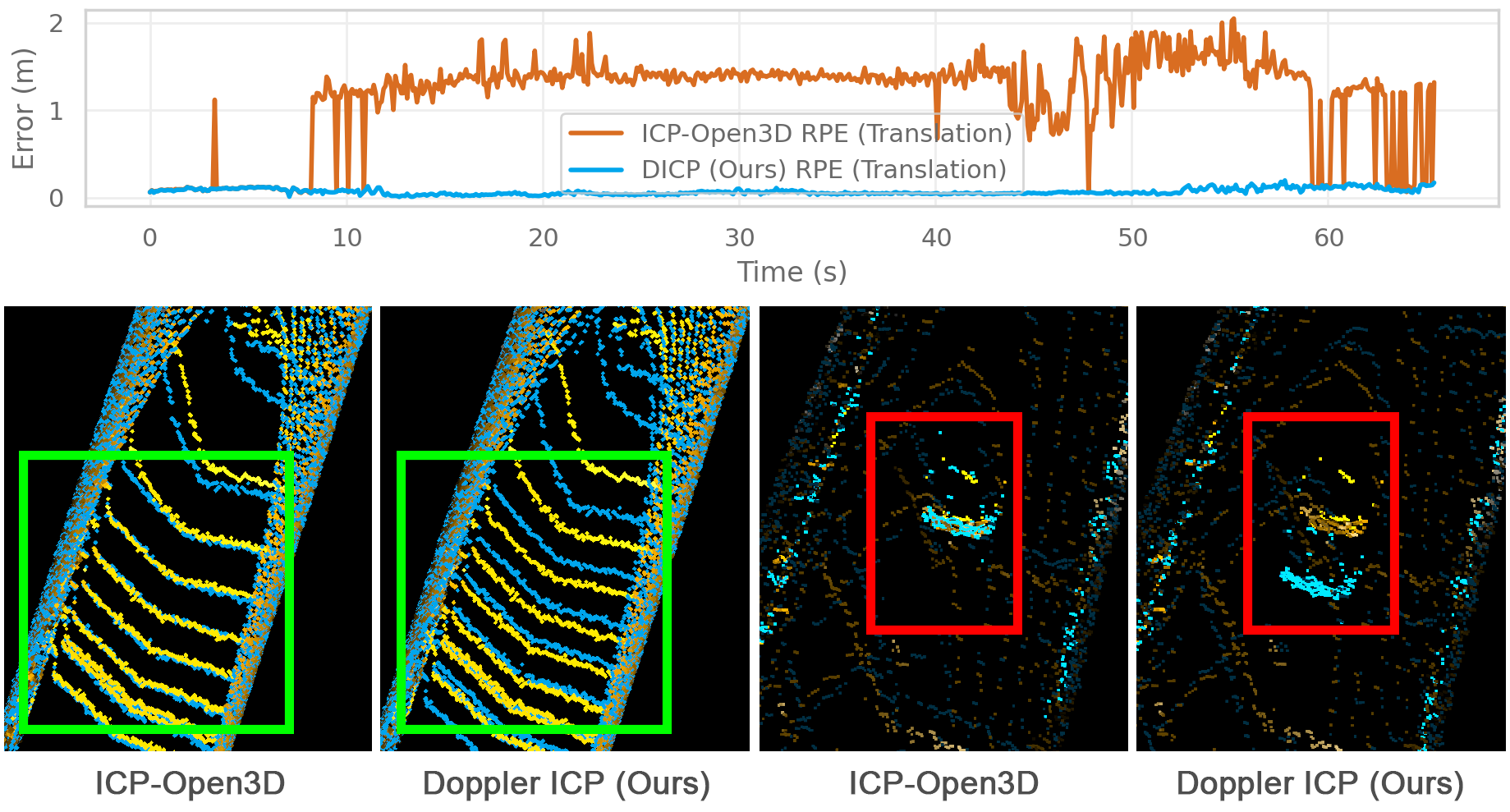}
  \caption{Plot showing the rise in RPE (translation) for \texttt{Baker-Barry Tunnel (Vehicles)} as the ego-vehicle enters it (top). ICP-Open3D registers to the scan pattern (left green-box), thereby estimating no motion; in another instance, it registers incorrectly to the motion of a vehicle in front (left red-box). Our method (DICP) estimates the motion of the ego-vehicle correctly in both failure cases.}
  \label{fig:icp-failure}
\end{figure}

\textbf{Better Pose Estimation:} As shown in Table \ref{tab:rmse-main} and Figure \ref{fig-traj}, our method outperforms the point-to-plane ICP method in feature-denied environments in terms of RPE and trajectory path error (the absolute difference between the accumulated distance estimated using ICP poses and using the ground-truth poses). In \texttt{Robin Williams Tunnel}, \texttt{Baker-Barry Tunnel (Empty)}, and the CARLA sequences, due to lack of geometric variations along planar surfaces, sequential point cloud scans appear very similar to each other, as if they were captured from the same location. This under-constraints the geometric objective function and leads to the failure of the point-to-plane ICP, as highlighted in the green boxes in Figure \ref{fig:icp-failure} and in Figure \ref{fig:reconstructions} (a), (b), and (f) as well. In scenarios where the point cloud geometry is dominated by dynamic objects (examples: \texttt{Brisbane Lagoon Freeway}, \texttt{Baker-Barry Tunnel (Vehicles)}, and \texttt{San Francisco City}), point-to-plane ICP registers to the motion of the moving object(s) instead of estimating the ego-vehicle motion correctly despite using robust kernels for outlier rejection, as shown in the red boxes in Figure \ref{fig:icp-failure} and white boxes in Figure \ref{fig:reconstructions} (c), (d), and (e). Our proposed method of pruning outlier correspondences makes it robust to such dynamic points.
Although we benchmark only against point-to-plane ICP, the results suggest that other geometric-based ICP variants are likely to fail due to the lack of geometric features in the dataset.

\textbf{Faster Convergence:} We observe that Doppler ICP converges faster, by an average factor of around 3.4 in terms of the number of iterations for convergence among the sequences in our \textit{No Seed Estimate} experiments (from Table \ref{tab:rmse-main}). Our method achieves similar relative pose errors for the \texttt{San Francisco City} sequence, which is rich in geometric features, but the faster convergence shows the benefit of using DICP. 

\subsection{Ablation Study}
\label{ablation}
\begin{figure}[h]
 \center
  \includegraphics[width=\columnwidth]{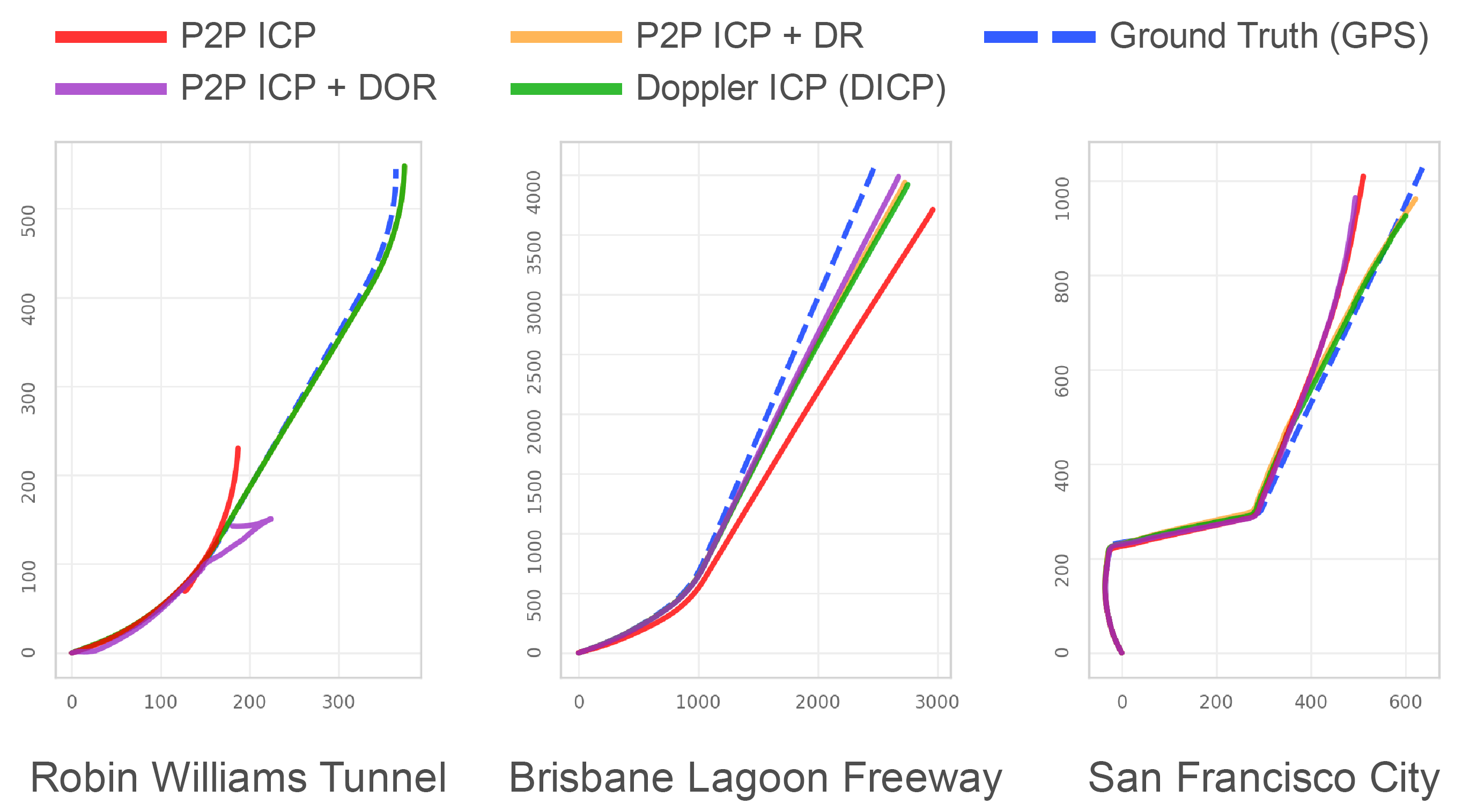}
  \caption{Trajectories of sequences from composition of poses estimated using  Open3D point-to-plane (P2P) ICP (red), point-to-plane ICP with dynamic outlier point rejection (purple), point-to-plane ICP with Doppler residuals (orange), and Doppler ICP which includes dynamic outlier point rejection (green). The ground truth trajectories (based on GPS) are shown in dashed blue lines.}
  \label{fig-ablation}
\end{figure}

We conduct additional experiments to investigate the individual benefits of the Doppler velocity residual term introduced in Section \ref{ss:doppler-residual}, and dynamic point outlier rejection introduced in Section \ref{ss:outlier-rejection}. The experimental setup and parameters remain the same as described in Section \ref{ss:implementation}. In order to not reject too many static point correspondences initially when the Doppler error between the prediction and the measurement is high, outlier rejection is only enabled after the second iteration of ICP. We use the estimated pose from the previous pair of registered scans under a constant-velocity motion-model assumption (similar to the \textit{With Seed Estimate} experiments from Section \ref{experiments}). We pick sequences from our dataset that have dynamic objects present in them to study the benefits of dynamic point rejection. The results are presented in Table \ref{tab:ablation} and the trajectories are shown in Figure \ref{fig-ablation}.

The relative pose error (RPE) for \texttt{Robin Williams Tunnel} and \texttt{Brisbane Lagoon Freeway} sequences reduces just by introducing the Doppler velocity residuals to the point-to-plane objective in ICP. Pruning possibly dynamic point correspondences alone on the \texttt{Brisbane Lagoon Freeway} sequence helps reduce the RPE and the error in total trajectory length. In \texttt{Robin Williams Tunnel}, pruning dynamic point correspondences does not help since the misalignment errors due to the lack of geometry constraints are already significant. We observed that, if the ICP algorithm incorrectly locks on to the pose of a dynamic object, enabling outlier rejection ends up removing the static points in the scene and thereby further diverging the solution from the ego-pose estimate. There is not a significant difference in the \texttt{San Francisco City} results due to the abundance of geometric features that sufficiently constrain the registration problem. We do, however, notice minor improvements in the total trajectory length estimated.

\begin{table}[t]
\caption{\label{tab:ablation}Ablation study results. \texttt{P2P} refers to the point-to-plane method from Open3D, \texttt{DR} refers to Doppler residuals, \texttt{DOR} indicates the usage of dynamic outlier rejection, and \texttt{DICP} refers to Doppler ICP that uses point-to-plane method, Doppler residuals, and outlier rejection.}
    \centering
    \begin{tabular}{llrrrrrrrr}
        \toprule
        \multicolumn{1}{c}{Sequence} &
        \multicolumn{1}{c}{Method} &
        \multicolumn{1}{c}{\shortstack{RPE\\Trans (m)}} & \multicolumn{1}{c}{\shortstack{RPE\\Rot (deg)}} & \multicolumn{1}{c}{\shortstack{Path\\Error (m)}} & \multicolumn{1}{c}{\shortstack{\# Iters\\(mean)}} \\
        \midrule
\multirow{4}{*}{\shortstack{Robin\\Williams\\Tunnel}}

& P2P
& {1.9343}  %
& {0.1533}  %
& {219.10}  %
& {20.3} \\  %

& P2P+DOR
& {2.6035}  %
& {0.6699}  %
& {298.04}  %
& \textbf{6.0} \\  %

& P2P+DR
& {0.0758}  %
& {0.1416}  %
& {0.05}  %
& {9.9} \\  %

& DICP
& \textbf{0.0758}  %
& \textbf{0.1415}  %
& \textbf{0.05}  %
& {9.8} \\  %

\midrule

\multirow{4}{*}{\shortstack{Brisbane\\Lagoon\\Freeway}}

& P2P
& {0.3270}  %
& {0.0915}  %
& {45.27}  %
& {10.6} \\  %

& P2P+DOR
& {0.1148}  %
& {0.0883}  %
& {9.67}  %
& {9.9} \\  %

& P2P+DR
& \textbf{0.1112}  %
& {0.0869}  %
& \textbf{4.38}  %
& {10.4} \\  %

& DICP
& {0.1301}  %
& \textbf{0.0869}  %
& {7.20}  %
& \textbf{9.7} \\  %

\midrule

\multirow{4}{*}{\shortstack{San\\Francisco\\City}}

& P2P
& {0.0323}  %
& \textbf{0.0479}  %
& {23.78}  %
& \textbf{5.9} \\  %

& P2P+DOR
& {0.0322}  %
& {0.0481}  %
& \textbf{4.55}  %
& {7.1} \\  %

& P2P+DR
& \textbf{0.0316}  %
& {0.0482}  %
& {10.18}  %
& {7.1} \\  %

& DICP
& {0.0317}  %
& {0.0482}  %
& {8.43}  %
& {6.0} \\  %

        \bottomrule
    \end{tabular}
\end{table}

\section{Conclusions}
\label{conclusions}
In this work, we presented a new algorithm for point cloud registration that leverages the Doppler velocity measurements from a point cloud captured by an FMCW range sensor. We provided a detailed formulation of a new Doppler velocity residual function and its joint optimization along with any of the existing geometric residuals commonly used in the ICP framework. Our experimental evaluation shows that this approach significantly improves the convergence rates and registration accuracy by providing additional optimization constraints, especially in feature-denied environments with non-distinctive and/or repetitive geometric surfaces where classical ICP variants tend to not converge properly. We also devised a method to reject dynamic points from the optimization process which otherwise would introduce errors in the estimated transform with existing ICP variants.

Despite the encouraging results, there are several avenues for future research to enhance the usage of our proposed method. We believe that the Doppler velocity objective function introduced in this paper, used in conjunction with other filtering or graph-based optimization techniques in a larger SLAM framework, could yield even better results, as well as fusing the data with other modalities of sensors such as cameras and IMUs. The DICP algorithm may be extended to additionally refine the extrinsic sensor calibration parameters or correct for the distortion due to motion in point cloud geometry. The presented algorithm could be augmented to incorporate data from the intensity channel from the sensor for a photometric error residual to further constrain the objective function. Furthermore, we believe that ensuring consistency between the current Doppler measurements and the previous Doppler measurement, in a scenario where there is not a significant amount of acceleration between measurements, one could leverage the fact that matched points with Doppler measurements have similar Doppler signatures to further constrain the optimization process.

\bibliographystyle{plainnat}

\end{document}